\documentclass[11pt,a4paper]{article}
\usepackage[hyperref]{emnlp-ijcnlp-2019}
\usepackage{times}
\usepackage{latexsym}

\usepackage{url}

\usepackage{enumitem}
\usepackage{graphicx}
\usepackage[labelformat=simple]{subcaption}

\usepackage{amsmath}
\usepackage{bm}
\usepackage[tone]{tipa}
\usepackage{cleveref}
\crefname{section}{§}{§§}
\Crefname{section}{Section}{}
\Crefname{figure}{Fig.}{}
\Crefname{algorithm}{Algorithm}{}
\Crefname{equation}{Equation}{}
\usepackage[ruled]{algorithm2e}
\usepackage{xcolor}
\usepackage{booktabs}
\usepackage{multirow}
\newcommand{\specialcell}[2][c]{%
  \begin{tabular}[#1]{@{}c@{}}#2\end{tabular}}
\usepackage{array}
\newcolumntype{L}[1]{>{\raggedright\let\newline\\\arraybackslash\hspace{0pt}}m{#1}}
\newcolumntype{C}[1]{>{\centering\let\newline\\\arraybackslash\hspace{0pt}}m{#1}}
\newcolumntype{R}[1]{>{\raggedleft\let\newline\\\arraybackslash\hspace{0pt}}m{#1}}
\DeclareMathOperator*{\argmax}{arg\,max}

\aclfinalcopy 

\title{Broad-Coverage Semantic Parsing as Transduction}

\author{Sheng Zhang\qquad Xutai Ma\qquad Kevin Duh\qquad Benjamin Van Durme\\
  Johns Hopkins University\\
  \texttt{\{zsheng2, xutai\_ma\}@jhu.edu}\\
  \texttt{\{kevinduh, vandurme\}@cs.jhu.edu}}

\date{}

\begin{document}
\maketitle
\begin{abstract}
    We unify different broad-coverage semantic parsing tasks 
    under a transduction  paradigm, 
    and propose an attention-based neural framework that 
    \emph{incrementally} builds a meaning representation via a sequence of semantic relations. 
    By leveraging multiple attention mechanisms, the  transducer can be 
    effectively trained without relying on a pre-trained aligner.
    Experiments conducted on three separate broad-coverage semantic parsing
    tasks -- AMR, SDP and UCCA --
    demonstrate that our attention-based neural transducer improves the 
    state of the art on both AMR 
    and UCCA, 
    and is competitive with the state of the art on SDP.

\end{abstract}

\section{Introduction}

Broad-coverage semantic parsing aims at mapping \emph{any} natural language text, 
regardless of its domain, genre, or even the language itself,
into a general-purpose meaning representation.
As a long-standing topic of interest in computational linguistics,
broad-coverage semantic parsing has targeted a number of 
meaning representation frameworks, including 
CCG~\citep{steedman1996surface,steedman2001syntactic}, DRS~\citep{Kamp1993-KAMFDT,bos-2008-wide},
AMR~\citep{AMR}, 
UCCA~\citep{ucca}, 
SDP~\citep{oepen-etal-2014-semeval,oepen-etal-2015-semeval}, 
and UDS~\citep{white-etal-2016-universal}.\footnote{Abbreviations respectively denote: Combinatory Categorical Grammar, Discourse Representation Theory, Abstract Meaning Representation,  Universal Conceptual Cognitive Annotation, Semantic Dependency Parsing, and Universal Decompositional Semantics.}
Each of these frameworks has their specific formal and linguistic assumptions.
Such framework-specific ``\emph{balkanization}''
results in a variety of framework-specific parsing approaches,
and the state-of-the-art semantic parser for one framework 
is not always applicable to another.
For instance, the state-of-the-art approaches to SDP parsing~\citep{P18-2077,P17-1186} 
are not directly transferable to AMR and UCCA because of 
the lack of explicit alignments between tokens in the sentence
and nodes in the semantic graph.

While transition-based approaches are adaptable to different broad-coverage 
semantic parsing tasks~\citep{AAAI1816549,hershcovich-etal-2018-multitask,E17-1051},
when it comes to representations such as AMR whose nodes are
\emph{unanchored} to tokens in the sentence, 
a pre-trained aligner has to be used to produce the reference transition sequences
\citep{camr,E17-1051,E17-1035}.
In contrast, there are attempts to develop attention-based approaches
in a graph-based parsing paradigm~\citep{P18-2077,sheng-etal-2019-amr}, 
but they lack parsing incrementality, which is advocated in terms of computational efficiency and cognitive modeling~\citep{Nivre:2004:IDD:1613148.1613156,Huang:2010:DPL:1858681.1858791}.

In this paper, we approach different broad-coverage 
semantic parsing tasks under a unified framework of transduction.
We propose an attention-based neural transducer
that extends the two-stage semantic parser of \citet{sheng-etal-2019-amr}
to directly transduce input text into a meaning representation in \emph{one} stage.
This transducer has properties of both transition-based approaches 
and graph-based approaches: 
on the one hand, it builds a meaning representation
incrementally via a sequence of semantic relations, similar to a transition-based parser; 
on the other hand, it leverages multiple attention mechanisms used in recent graph-based parsers,
thereby removing the need for pre-trained aligners.

Requiring only minor task-specific adaptations, we apply this framework
to three separate broad-coverage semantic parsing tasks: AMR, SDP, and UCCA.
Experimental results show that 
our neural transducer outperforms the state-of-the-art parsers 
on AMR (77.0\% F1 on LDC2017T10 and 71.3\% F1 on LDC2014T12) and 
UCCA (76.6\% F1 on the English-Wiki dataset v1.2),
and is competitive with the state of the art on SDP 
(92.2\% F1 on the English DELPH-IN MRS dataset). 

\section{Background and Related Work}

We provide summary background on the meaning representations we target,
and review related work on parsing for each.

\begin{figure*}[!ht]
\centering
\includegraphics[width=0.96\textwidth]{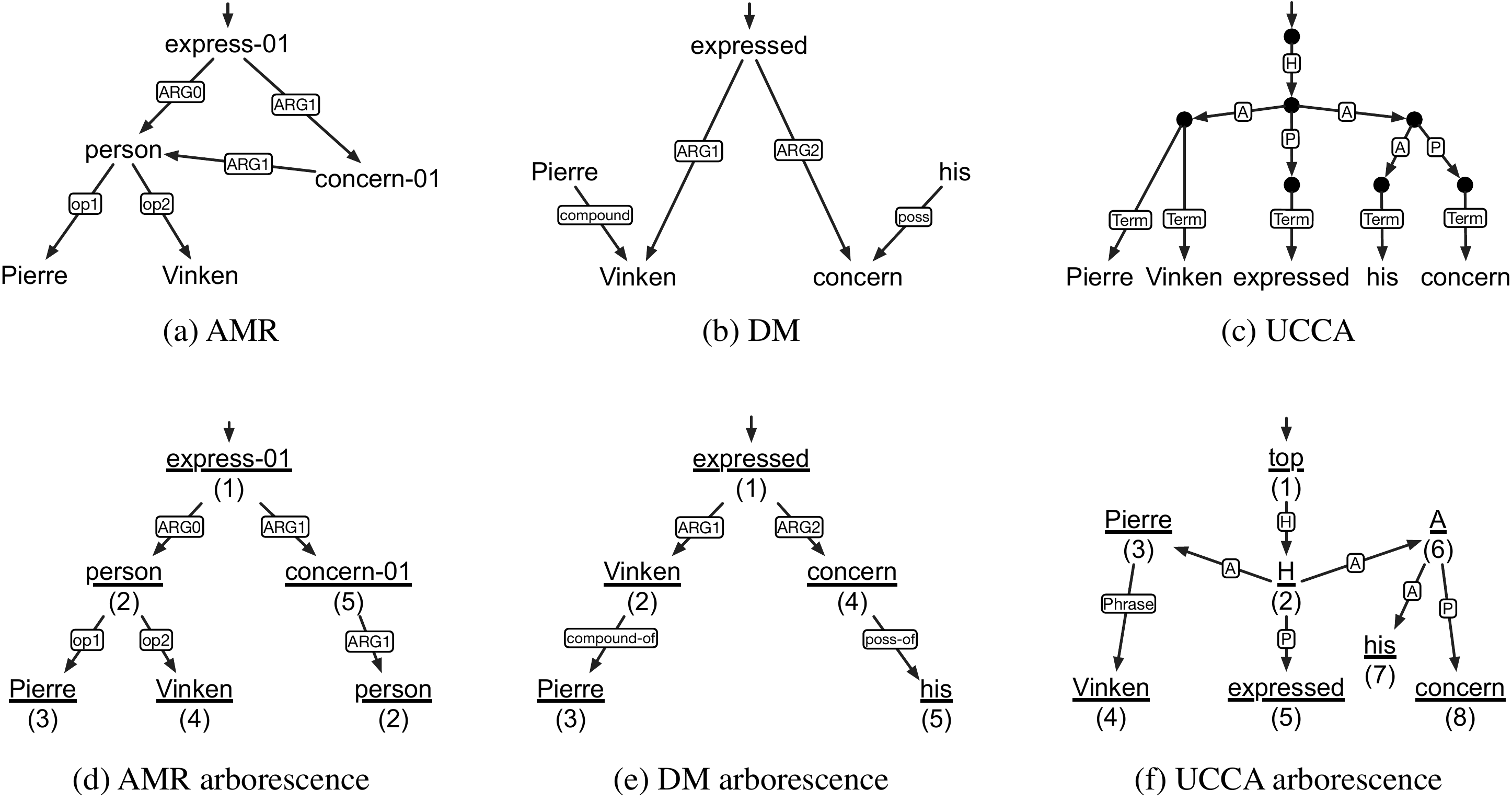}
\caption{Meaning representation in the task-specific format -- 
(a) AMR, (b) DM, and (c) UCCA --
for an example sentence ``\emph{Pierre Vinken expressed his concern}''.
Meaning representation (d), (e) and (f) are in the unified arborescence format,
which are converted from (a), (b) and (c) respectively.
\label{fig:semantic-graphs}}
\end{figure*}

\noindent\textbf{Abstract Meaning Representation}
(AMR; \citealp{AMR}) encodes sentence-level
semantics, such as predicate-argument information, reentrancies, named entities,
negation and modality, into a rooted, directed, and usually acyclic graph 
with node and edge labels. 
AMR graphs abstract away from syntactic realizations, i.e., 
there is no explicit correspondence between elements of the graph and the surface utterance. 
\Cref{fig:semantic-graphs}(a) shows an example AMR graph.

Since its first general release in 2014, AMR has been a popular
target of data-driven semantic parsing, notably in two SemEval shared tasks
\citep{may-2016-semeval,may-priyadarshi-2017-semeval}.
Graph-based parsers build AMRs by identifying concepts and scoring edges
between them, either in a pipeline~\citep{jamr}, 
or jointly~\citep{D16-1065,P18-1037,sheng-etal-2019-amr}. 
This two-stage parsing process limits the parser incrementality.
Transition-based parsers either transform dependency trees into AMRs~\citep{camr,S16-1181,S16-1180},
or employ transition systems specifically tailored to AMR parsing
\citep{E17-1051,ballesteros-al-onaizan-2017-amr}.
Transition-based parsers rely on pre-trained aligner produce the reference transitions.
Grammar-based parsers leverage external semantic resources to 
derive AMRs compositionally based on CCG rules~\citep{D15-1198}, 
or SHRG rules~\citep{peng-etal-2015-synchronous}.
Another line of work uses neural model translation models to convert sentences
into \emph{linearized} AMRs~\citep{barzdins-gosko-2016-riga,E17-1035}, 
but has relied on data augmentation to produce effective parsers~\citep{amr-seq2seq,P17-1014}.
Our parser differs from the previous ones in that it has incrementality without relying
on pre-trained aligners, and
can be effectively trained without data augmentation. 

\noindent\textbf{Semantic Dependency Parsing}
(SDP) was introduced
in 2014 and 2015 SemEval shared tasks~\citep{oepen-etal-2014-semeval,oepen-etal-2015-semeval}.
It is centered around three semantic formalisms --
DM (DELPH-IN MRS; \citealp{flickinger-2012-deepbank,oepen-lonning-2006-discriminant}),
PAS (Predicate-Argument Structures; \citealp{miyao-tsujii-2004-deep}),
and PSD (Prague Semantic Dependencies; \citealp{hajic-etal-2012-announcing}) -- representing
predicate-argument relations between content words in a sentence.
Their annotations have been converted into bi-lexical dependencies,
forming directed graphs whose nodes injectively correspond to surface lexical units,
and edges represent semantic relations between nodes.
In this work, we focus on only the DM formalism.
\Cref{fig:semantic-graphs}(b) shows an example DM graph.

Most recent parsers for SDP are graph-based: \citet{P17-1186,peng-etal-2018-learning} 
use a max-margin classifier on top of a BiLSTM, with the factored score for each graph over predicates, 
unlabeled arcs, and arc labels.
Multi-task learning approaches and disjoint data have been used to improve the parser performance.
\citet{P18-2077} extend an LSTM-based syntactic dependency parser to produce
graph-structured dependencies, and carefully tune it to state of the art performance. 
\citet{AAAI1816549} extend the transition system of \citet{choi-mccallum-2013-transition} 
to produce non-projective trees,
and use improved versions of stack-LSTMs~\citep{dyer-etal-2015-transition}
to learn representation for key components.
All of these are specialized for bi-lexical dependency parsing, 
whereas our parser can effectively produce both bi-lexical semantics graphs, 
and graphs that are less anchored to the surface utterance.

\noindent\textbf{Universal Conceptual Cognitive Annotation}
(UCCA; \citealp{ucca}) targets
a level of semantic granularity that abstracts away from syntactic paraphrases
in a typologically-motivated, cross-linguistic fashion.
Sentence representations in UCCA are directed acyclic graphs (DAG), 
where terminal nodes correspond to surface lexical tokens, 
and non-terminal nodes to semantic units that participate in super-ordinate relations.
Edges are labeled, indicating the role of a child in the relation the parent represents. 
\Cref{fig:semantic-graphs}(c) shows an example UCCA DAG.

The first UCCA parser is proposed by \citet{hershcovich-etal-2017-transition},
where they extend a transition system to produce DAGs.
To leverage other semantic resources,
\citet{hershcovich-etal-2018-multitask} is one of the few attempts to present
(lossy) conversion from AMR, SDP and Universal Dependencies (UD; \citealp{ud}) 
to a unified UCCA-based DAG format. 
They explore multi-task learning under the unified format.
While multi-task learning improves UCCA parsing results, 
it shows poor performance on AMR, SDP and UD parsing.
In contrast, different semantic parsing tasks 
are formalized in our unified transduction paradigm with no loss,
and our approach achieves
state-of-the-art or competitive performance on each task,  using only single-task data.

\section{Unified Transduction Problem}

\subsection{Unified Arborescence Format}
\label{sec:unified-format}

We first introduce a unified target format for different broad-coverage semantic parsing tasks.
Meaning representation in the unified format is an arborescence (aka, a directed rooted tree), 
which is converted from its corresponding task-specific semantic graph 
via the following \emph{reversible} steps:

\noindent\textbf{AMR}
Reentrancy is what can make an AMR graph not an arborescence (it introduces cycles).
Following \citet{sheng-etal-2019-amr},
we convert an AMR graph into an arborescence by duplicating nodes that
have reentrant relations; that is, whenever a node has a reentrant relation,
we make a copy of that node and use the copy to participate in the relation,
thereby resulting in an arborescence.
Next, in order to preserve the reentrancy information, we assign a node index to each node.
Duplicated nodes are assigned the same index as the original node.
\Cref{fig:semantic-graphs}(d) shows an AMR arborescence converted from \Cref{fig:semantic-graphs}(a):
two ``\emph{person}'' nodes have the same node index 2.
The original AMR graph can be recovered by merging identically indexed nodes.

\noindent\textbf{DM}
We first break the DM graph into a set of \emph{weakly} connected subgraphs.
For each subgraph, if it has the \emph{top} node, we treat \emph{top} as root; 
otherwise, we treat the node with the max outdegree as root.
We then run depth-first traversal over each subgraph from its root to yield an arborescence,
and repeat the following three steps until no more edges can be added to the arborescence:
(1) we run breadth-first traversal over the arborescence from the root until
we find a node that has an incoming edge not belonging to the arborescence;
(2) we reverse the edge and add a \texttt{-of} suffix to the edge label;
(3) we run depth-first search from that node to include more edges to the arborescence.
During the whole process, we add node indices and duplicate reentrant nodes in the same way
as AMR conversion.
Finally, we connect arborescences by adding a \texttt{null} edge from \emph{top} to other arborescence roots.
\Cref{fig:semantic-graphs}(e) shows a DM arborescence converted from \Cref{fig:semantic-graphs}(b).
The original DM graph can be recovered by removing \texttt{null} edges,
merging identically indexed nodes, and reversing edges with \texttt{-of} suffix.

\begin{figure*}[!ht]
\centering
\includegraphics[width=0.98\textwidth]{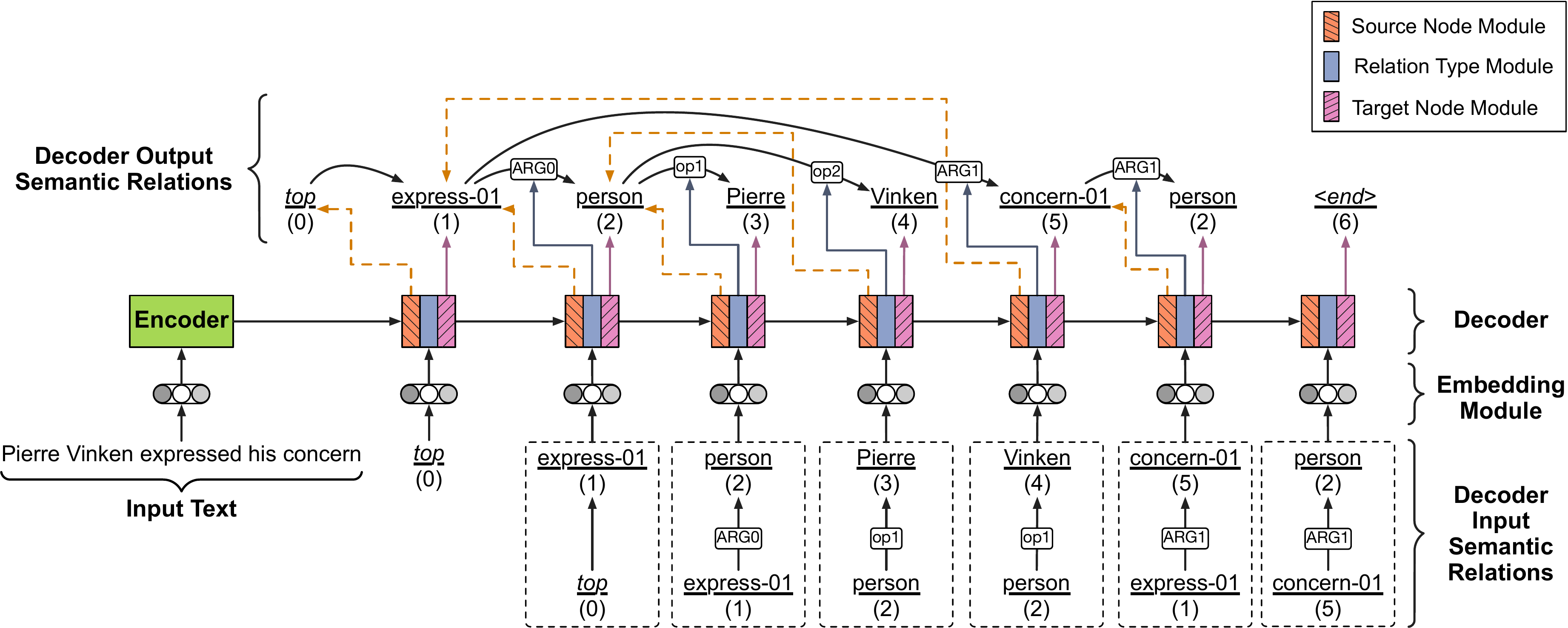}
\caption{The encoder-decoder architecture of our attention-based neural transducer.
An encoder encodes the input text into hidden states.
A decoder is composed by three modules: a target node module, a relation type module,
and a source node module.
At each decoding time step, the decoder takes the previous semantic relation as input, and 
outputs a new semantic relation in a \emph{factorized} way: firstly, the target node module produces a new target node; 
secondly, the source node module \emph{points} to a preceding node as a new source node;
finally, the relation type module predicts the relation type between source and target nodes.
\label{fig:model}}
\end{figure*}

\noindent\textbf{UCCA}
To date, official UCCA evaluation only considers UCCA's \emph{foundational} layer,
which is already an arborescence.
We convert it to the unified arborescence format by first collapsing subgraphs of pre-terminal nodes:
we replace each pre-terminal node with its first terminal node; 
if the pre-terminal node has other terminals, we add a special \texttt{phrase} edge from 
the first terminal node to other terminal nodes.
The collapsing step largely reduces the number of terminal nodes in UCCA.
We then add labels to the remaining non-terminal nodes.
Each node label is simply the same as its incoming edge label. We find that adding node labels improves performance of our neural transducer
(See \Cref{sec:results} for the experimental results).
Lastly, we add node indices in the same way as AMR conversion.
\Cref{fig:semantic-graphs}(f) shows a DM arborescence converted from \Cref{fig:semantic-graphs}(c).
The original UCCA DAG can be recovered by expanding pre-terminal subgraphs,
and removing non-terminal node labels.

\subsection{Problem Formalization}

For any broad-coverage semantic parsing task, we denote the input text by $X$,
and the output meaning representation in the unified arborescence format by $Y$, where $X$ is a sequence of tokens $\langle x_1,x_2, ..., x_n\rangle$ and
$Y$ can be decomposed as
a sequence of semantic relations $\langle y_1, y_2, ..., y_m\rangle$.
A relation $y$ is a tuple $\langle u, d^{u}, r, v, d^v\rangle$, consisting of a source node label $u$, a source node index $d^u$, a relation type $r$, a target node label $v$, and a target node index $d^v$.

Let $\mathcal{Y}$ be the \emph{output space}.
The unified transduction problem is 
to seek the most-likely sequence of semantic relations $\hat{Y}$
given $X$:
\begin{align*}
    \hat{Y} & = \argmax_{Y\in\mathcal{Y}}\text{P}(Y\mid X)\\
            & = \argmax_{Y\in\mathcal{Y}}\prod_i^m\text{P}(y_i\mid y_{<i}, X)
\end{align*}

\section{Transducer}
To tackle the unified transduction problem,
we introduce an attention-based neural transducer
that extends \citet{sheng-etal-2019-amr}'s attention-based parser.
Their attention-based parser addresses semantic parsing in a two-stage process:
it first employs an extended variant of pointer-generator network~\citep{pointer-generator} to convert
the input text into a list of nodes, and then uses a deep biaffine
graph-based parser~\citep{dozat2016deep} with a maximum spanning tree (MST) algorithm to create edges.
In contrast, our attention-based neural transducer directly transduces
the input text into a meaning representation in \emph{one} stage via a sequence of semantic relations.
A high-level model architecture of our transducer is depicted in \Cref{fig:model}:
an \emph{encoder} first encodes the input text into hidden states;
and then conditioned on the hidden states, 
at each decoding time step, a \emph{decoder} takes the previous semantic relation as input,
and outputs a new semantic relation, which includes a target node, a relation type, and a source node.

Specifically, there a significant difference between \citet{sheng-etal-2019-amr} and our model:
\citet{sheng-etal-2019-amr} first predicts nodes, and then edges. 
These two stages are done \emph{separately} (except that a shared encoder is used). 
At the node prediction stage, their model has no knowledge of edges, and therefore node prediction is performed purely based previous nodes.
At the edge prediction stage, their model predicts the head of each node in parallel.
Head prediction of one node has no constrains or impact on another.
As a result, MST algorithms have to be used to search for a valid prediction.
In comparison, our model does not have two separate stages for node and edge prediction.
At each decoding step, our model predicts not only a node, but also the incoming edge to the node, 
which includes a source and a relation type. See \Cref{fig:model} for an example.  
The predicted node and incoming edge together with previous predictions form a partial semantic graph, 
which is used as input of the next decoding step for the next node and incoming edge prediction.
Our model therefore makes predictions based on the partial semantic graph, which helps prune the output space for both nodes and edges.
Since at each decoding step, we assume the incoming edge is always from a preceding node (see \Cref{sec:training} for the details),
the predicted semantic graph is guaranteed to be a valid arborescence, and a MST algorithm is no longer needed.

\subsection{Encoder}

At the encoding stage, we 
employ an encoder embedding module to convert the input text
into vector representations, and a BiLSTM is used to encode vector representations into hidden states.

\noindent\textbf{Encoder Embedding Module} concatenates word-level embeddings from
GloVe~\citep{glove} and BERT\footnote{
We use average pooling in the same way as \citet{sheng-etal-2019-amr} to get word-level embeddings from BERT.}
\citep{devlin2018bert}, 
char-level embeddings from CharCNN~\citep{charCNN}, 
and randomly initialized embeddings for POS tags.

For AMR, it includes extra randomly initialized embeddings for anonymization indicators 
that tell the encoder whether a token is an anonymized token from preprocessing. 

For UCCA, it includes extra randomly initialized embeddings for NER tags, syntactic dependency labels, 
punctuation indicators, and shapes that are provided in the UCCA official dataset.

\noindent\textbf{Multi-layer BiLSTM}~\cite{hochreiter1997long} is defined as: 
\begin{equation}
\mathbf{s}^l_t = 
\left[
    \begin{array}{ll}
         \overrightarrow{\mathbf{s}}^l_t  \\
         \overleftarrow{\mathbf{s}}^l_t  \\
    \end{array}
    \right] = 
\left[
    \begin{array}{ll}
        \overrightarrow{\textsc{lstm}}(\mathbf{s}^{l-1}_t, \mathbf{s}^l_{t-1})\\
        \overleftarrow{\textsc{lstm}}(\mathbf{s}^{l-1}_t, \mathbf{s}^l_{t+1})
    \end{array}
    \right],
\end{equation}
where $\mathbf{s}^l_t$ is the $l$-th layer hidden state at time step $t$;
$\mathbf{s}^t_i$ is the embedding module output for token $x_t$.

\subsection{Decoder}

\textbf{Decoder Embedding Module}
at decoding time step $i$ converts 
elements in the input semantic relation $\langle u_i, d^u_i, r_i, v_i, d^v_i\rangle$
into 
vector representations $\langle\mathbf{u}_i, \mathbf{d}^u_i, \mathbf{r}_i, \mathbf{v}_i, \mathbf{d}^v_i\rangle$:\footnote{While training, the input semantic relation is from the reference sequence of relations; 
at test time, it is the previous decoder output semantic relation.}

$\mathbf{u}_i$ and $\mathbf{v}_i$ are concatenations of word-level embeddings from GloVe, 
char-level embeddings from CharCNN, and randomly initialized embeddings for POS tags.
POS tags for source and target nodes are inferred at runtime:
if a node is copied from input text, the POS tag of the corresponding token is used; if it is copied from a preceding node, the POS tag of the preceding node is used; otherwise, an \texttt{UNK} tag is used.

$\mathbf{d}^u_i, \mathbf{d}^v_i$ and $\mathbf{r}_i$ are randomly initialized embeddings
for source node index, target node index, and relation type.

Next, the decoder outputs a new semantic relation in a \emph{factorized} way depicted in \Cref{fig:model}: 
First, a target node module takes vector representations of the previous semantic relation, 
and predicts a target node label as well as its index. 
Then, a source node module predicts a source node via \emph{pointing} to a preceding node. 
Lastly, a relation type module takes the predicted source and target nodes,
and predicts the relation type between them. 

\noindent\textbf{Target Node Module} converts vector representations of the input semantic relation 
into a hidden state $\mathbf{z}_i$ in the following way:

\begin{gather}
    \mathbf{z}_i = \textsc{ffn}^{(\text{relation})}([\mathbf{h}^l_i; \mathbf{c}_i;\mathbf{r}_i;\mathbf{u}_i;\mathbf{d}^u_i])\\
    \mathbf{h}^l_i = \textsc{lstm}(\mathbf{h}^{l-1}_i, \mathbf{h}^l_{i-1})\label{eq:lstm}\\
    \textsc{ffn}(\mathbf{x}) = \mathbf{W}\mathbf{x}+\mathbf{b}
\end{gather}
where an $l$-layer LSTM generates contextual representation $\mathbf{h}^l_i$ for \emph{target} node $v_i$
(for initialization, $\mathbf{h}^0_i = [\mathbf{v}_i;\mathbf{d}^v_i]$,
$\mathbf{h}^l_0=[\overleftarrow{\mathbf{s}}^l_1;\overrightarrow{\mathbf{s}}^l_n]$).
A feed-forward neural network
$\textsc{ffn}^{(\text{relation})}$ generates 
the hidden state $\mathbf{z}_i$ of the input semantic relation by combining
contextual representation $\mathbf{h}^l_i$ for target node $v_i$,
encoder context vector $\mathbf{c}_i$,
and vector representations $\mathbf{r}_i, \mathbf{u}_i, \mathbf{d}^u_i$ for relation type $r_i$, 
source node label $u_i$ and source node index $d^u_i$.

Encoder context vector $\mathbf{c}_i$ is a weighted-sum of encoder hidden states $\mathbf{s}^l_{1:n}$. 
The weight is attention $\mathbf{a}^{(\text{enc})}_i$ from the decoder at decoding step $i$ to encoder hidden states:
\begin{gather}
    \mathbf{a}^{(\text{enc})}_i = \textrm{softmax}\big(\textsc{mlp}^{(\text{enc})}([\mathbf{h}^l_i; \mathbf{s}^l_{1:n}])\big)\\
    \textsc{mlp}(\mathbf{x}) = \textsc{elu}(\mathbf{W}\mathbf{x}+\mathbf{b})
\end{gather}

Given the hidden state $\mathbf{z}_i$ for input semantic relation, we use an extended 
variant of pointer-generator network to compute 
the probability distribution of next target node label $v_{i+1}$:
\begin{gather}
    \text{P}(v_{i+1}) =  p_\text{gen}\mathbf{p}^{(\text{vocab})}_i\oplus
    p_\text{enc} \mathbf{a}^{(\text{enc})}_i\oplus
    p_\text{dec} \mathbf{a}^{(\text{dec})}_i\\
    \mathbf{p}^{(\text{vocab})}_i = \textrm{softmax}\big(\textsc{ffn}^{(\text{vocab})}(\mathbf{z}_i)\big)\\
    \mathbf{a}^{(\text{dec})}_i = \textrm{softmax}\big(\textsc{mlp}^{(\text{dec})}([\mathbf{z}_i; \mathbf{z}_{1:i-1}])\big)\\
     [p_\textrm{gen},p_\textrm{enc},p_\textrm{dec}] = \textrm{softmax}\big(\textsc{ffn}^{(\textrm{switch})}(\mathbf{z}_i)\big)
\end{gather}
$\text{P}(v_{i+1})$ is a hybrid of three parts:
(1) emitting a new node label from a pre-defined vocabulary via probability distribution $\mathbf{p}^{(\text{vocab})}_i$;
(2) copying a token from the encoder input text as node label via encoder-side attention $\mathbf{a}^{(\text{enc})}_i$;
and (3) copying a node label from preceding target nodes via decoder-side attention $\mathbf{a}^{(\text{dec})}_i$.
Scalars $p_\textrm{gen},p_\textrm{enc}$ and $p_\textrm{dec}$ act as a soft switch to control the production
of target node label from different sources.

The next target node index $d^v_{i+1}$ is assigned based on the following rule:
\begin{align*}
    d^v_{i+1} = \left \{
        \begin{aligned}
            d^v_j, &\ \textrm{if}\ v_{i + 1} \textrm{ copies its antecedent } v_j\textrm{.}\\
            i + 1, &\ \textrm{otherwise}.
        \end{aligned}
    \right.
\end{align*}

\noindent\textbf{Source Node Module} produces the next source node label $u_{i + 1}$ via
\emph{pointing} to a node label among preceding \emph{target} node labels (the dotted arrows shown in \Cref{fig:model}).
The probability distribution of next source node label $u_{i+1}$ is defined as 
\begin{equation}
    \text{P}(u_{i+1}) = \textrm{softmax}\big(\textsc{biaffine}(\mathbf{h}^{(\text{start})}_{i+1},
    \mathbf{h}^{(\text{end})}_{1:i})\big)\\
\end{equation}
where \textsc{biaffine} is a biaffine function~\citep{dozat2016deep}.
$\mathbf{h}^{(\text{start})}_{i+1}$ is the vector representation for the \emph{start} of the pointer. 
$\mathbf{h}^{(\text{end})}_{1:i}$ are vector representations for possible \emph{ends} of the pointer. 
They are computed by two multi-layer perceptrons: 
\begin{gather}
\mathbf{h}^{(\text{start})}_{i+1} = \textsc{mlp}^{(\text{start})}(\mathbf{h}^l_{i+1})\\
\mathbf{h}^{(\text{end})}_{1:i} = \textsc{mlp}^{(\text{end})}(\mathbf{h}^l_{1:i})
\end{gather}

Note that $\mathbf{h}^l_{i+1}$ is the LSTM hidden state for target node $v_{i+1}$, 
generated by \Cref{eq:lstm} in the target node module.
We reuse LSTM hidden states from the target node module such that we can train the decoder modules jointly.

Then, the next source node index $d^u_{i+1}$ is the same as the target node the module points to.

\noindent\textbf{Relation Type Module} also reuses LSTM hidden states from the target node module
to compute the probability distribution of next relation type $r_{i+1}$.
Assuming that the source node module points to target node label $v_j$ as the next source node label,
The next relation type probability distribution is computed by:
\begin{gather}
\text{P}(r_{i+1}) = \text{softmax}\big(\textsc{bilinear}(\mathbf{h}^{(\text{rel-src})}_{i+1}, \mathbf{h}^{(\text{rel-tgt})}_{i+1})\big)\\
\mathbf{h}^{(\text{rel-src})}_{i+1} = \textsc{mlp}^{(\text{rel-src})}(\mathbf{h}^l_{j})\\
\mathbf{h}^{(\text{rel-tgt})}_{i+1} = \textsc{mlp}^{(\text{rel-tgt})}(\mathbf{h}^l_{i+1})
\end{gather}

\subsection{Training}
\label{sec:training}

To ensure that at each decoding step, the source node can be found in the preceding nodes, 
we create the reference sequence of semantic relations by running a \emph{pre-order} traversal over the reference arborescence.
The pre-order traversal only determines the order between a node and its children.
As for the order of its children, we sort them in alphanumerical order in the case of AMR, following \citet{sheng-etal-2019-amr}.
In the case of SDP, we sort the children based on their order in the input text.
In the case of UCCA, we sort the children based on their UCCA node ID. 

Given a training pair $\langle X, Y\rangle$, the optimization objective is to maximize 
the decomposed conditional log likelihood $\sum_i\log\big(\text{P}(y_i\mid y_{<i}, X)\big)$,
which is approximated by:
\begin{equation}
    \sum_i \log\big(\text{P}(u_i)\big)+\log\big(\text{P}(r_i)\big)+\log\big(\text{P}(v_i)\big)
\end{equation}

We also employ label smoothing~\cite{szegedy2016rethinking} 
to prevent overfitting, 
and include a coverage loss~\citep{pointer-generator} to penalize repetitive nodes:
$\textrm{covloss}_i=\sum_t\textrm{min}(\mathbf{a}^{(\textrm{enc})}_i[t],\textbf{cov}_i[t])$, where
$\textbf{cov}^i = \sum_{j=0}^{i-1}\mathbf{a}^{(\text{enc})}_j$.

\subsection{Prediction}

Our transducer at each decoding time step looks for the source node  
from the preceding nodes, which ensures that the output of a greedy
search is already a valid arborescence $\hat{Y}$:
\begin{equation*}
    \text{P}(\hat{Y}\mid X)=\prod_i\max_{u_i}\text{P}(u_i)\max_{r_i}
    \text{P}(r_i)\max_{v_i}\text{P}(v_i)
\end{equation*}
Therefore, a MST algorithm such as the Chu-Liu-Edmonds algorithm at $\mathcal{O}(EV)$ used in \citet{sheng-etal-2019-amr}  
is no longer needed,\footnote{$E$ denotes the number of edges. $V$ the number of nodes.}
and the decoding speed of our transducer is $\mathcal{O}(V)$.
Moreover, since our transducer builds the meaning representation 
via a sequence of semantic relations,
we implement a beam search over relation in Algo. \ref{algo:beamsearch}.
Compared to the beam search of \citet{sheng-etal-2019-amr} that only
returns top-$k$ nodes,
our beam search finds the top-$k$ relation scores,
which includes source nodes, relation types and target nodes.

\begin{algorithm}[!ht]
\SetAlCapNameFnt{\small}
\SetAlCapFnt{\small}
\small
\SetKwData{Bucket}{\textsc{bucket}}
\SetKwData{Beam}{beam}
\SetKwData{NewBeam}{new\_beam}
\SetKwInOut{Input}{Input}
\SetKwInOut{Output}{Output}

\Input{The input text $X$.}
\Output{A sequence of relations $Y=\{y_1,...y_m\}$.}
\tcp{Initialization.}
$i, \text{score}\leftarrow 0, 0$\;
$Y, \texttt{finished}\leftarrow \{\}, \{\}$\;
\texttt{beam}$\leftarrow \{ \{Y, \text{score}\}\}$\;
\BlankLine
\tcp{Encoding.}
\texttt{encode($X$)}\;
\BlankLine
\tcp{Decoding.}
\For{$i\leftarrow 1$ \KwTo \text{\upshape MaxLength}}{
    \texttt{new\_beam} $\leftarrow \{\}$\;
    $\{Y, \text{score}\}$ = \texttt{beam.pop()}\;
    \For{$v_i$ \text{\upshape in} \texttt{\upshape topK}($\text{\upshape P}(v_i)$)}{
        \uIf{$v_i$ = \texttt{\upshape EOS}}{
            \texttt{finished.push}($\{Y, \text{score}\})$\;
        }\Else{
            \For{$u_i\leftarrow v_0$ \KwTo $v_{i-1}$}{
                \For{$r_i$ \text{\upshape in RelationTypeSet}}{
                    $Y \leftarrow Y\cup\{\langle u_i, r_i, v_i\rangle\}$\;
                    $\text{score} \leftarrow \text{score} + \log\big(\text{P}(u_i)\big) + \log\big(\text{P}(r_i)\big) + \log\big(\text{P}(v_i)\big)$\;
                    \texttt{new\_beam.push}(\{Y, \text{score}\})\;
                }
            }
        }
    }
    \texttt{beam} $\leftarrow$ \texttt{new\_beam.topK}()\; 
}
\BlankLine
\tcp{Finishing.}
\While{\texttt{\upshape beam.not\_empty()}}{
    $\{Y, \text{score}\}$ $\leftarrow$ \texttt{beam.pop()}\;
    \texttt{finished.push}($\{Y, \text{score}\})$\;
}
$\{Y, \text{score}\}$ $\leftarrow$ \texttt{finished.topK(k=1)}\;
\BlankLine
\Return $Y$\;
\caption{Beam Search over Semantic Relations.\label{algo:beamsearch}}
\end{algorithm}

\section{Data Pre- and Post-processing}
\noindent\textbf{AMR} Pre- and post-processing steps are similar to
those of \citet{sheng-etal-2019-amr}:
in preprocessing, we anonymize subgraphs of entities, remove senses, 
and convert resultant AMR graphs into the unified format;
in post-processing, we assign the most frequent sense for nodes, 
restore Wikipedia links using the DBpedia Spotlight API~\citep{isem2013daiber},
add polarity attributes based on rules observed from training data,
and recover the original AMR format from the unified format.

\noindent\textbf{DM} No pre- or post-processing is done to DM
except converting them into the unified format, 
and recovering them from predictions.

\noindent\textbf{UCCA} During training, multi-sentence input text 
and its corresponding DAG are split into 
single-sentence training pairs based on rules observed 
from training data.
At test time, we split multi-sentence input text, 
and join the predicted graphs into one.
We also convert the original format to the unified format in preprocessing, 
and recover the original DAG format in post-processing.

\begin{table}[!ht]
\small
\centering
\begin{tabular}{ l c }
\toprule
\multicolumn{2}{c}{\textbf{Hidden Size}}\\ 
Glove                                      &300              \\ 
BERT                                     &1024              \\ 
POS / NER / Dep / Shapes &100 \\
\noindent\begin{tabular}{@{}l}
Anonymization / Node index\\
\end{tabular} &50 \\ 
CharCNN kernel size &3 \\
CharCNN channel size &100 \\
Encoder &2@512 \\
Decoder &2@1024 \\
Biaffine input size  &256                                \\
Bilinear input size  &\begin{tabular}{lc}AMR &128 \\ DM & 256   \\ UCCA&128  \end{tabular}                                  \\ \midrule
\multicolumn{2}{c}{\textbf{Optimizer}} \\
Type                                    &ADAM             \\
Learning rate                          &0.001           \\
Maximum gradient norm                         &5.0              \\
Coverage loss weight $\lambda$ &1.0              \\
Label smoothing $\epsilon$     &0.1              \\
Beam size                      &5                \\
Batch size                     &64           \\
Dropout rate &\begin{tabular}{lc}AMR &0.33 \\ DM & 0.2   \\ UCCA&0.33  \end{tabular} \\ \midrule
\multicolumn{2}{c}{\textbf{Vocabulary}}                                                    \\
Encoder-side vocab size                 & \begin{tabular}{lc}AMR 1.0 &9200 \\AMR 2.0 &18000   \\ DM & 11000   \\ UCCA&10000  \end{tabular}  \\
Decoder-side vocab size& \begin{tabular}{lc}AMR 1.0 &7300 \\AMR 2.0 &12200    \\ DM & 11000   \\ UCCA&10000  \end{tabular}  \\ \bottomrule
\end{tabular}
\caption{Hyperparameter settings}
\label{tab:hyper}
\end{table}

\section{Experiments}

\subsection{Data and Setup}
We evaluate our approach on three separate broad-coverage semantic parsing tasks:
(1) AMR 2.0 (LDC2017T10) and 1.0 (LDC2014T12);
(2) the English DM dataset from SemEval 2015 Task 18 (LDC2016T10); 
(3) the UCCA English Wikipedia Corpus v1.2~\citep{ucca,hershcovich-etal-2019-semeval}.
The train/dev/test split follows the official setup.
Our model is trained on two GeForce GTX TITAN X GPUs with early stop based on the dev set.
We fix BERT parameters similar to \citet{sheng-etal-2019-amr}
due to the limited GPU memory.
Hyperparameter setting for each task is provided in \Cref{tab:hyper}.

\subsection{Results}
\label{sec:results}

\begin{table}[!ht]
\centering
\begin{tabular}{@{}cll@{}}
\toprule
Data                     & Parser & \multicolumn{1}{c}{F1(\%)} \\ \midrule
\multirow{5}{*}{\specialcell{AMR\\ 2.0}}
                         & \citet{cai-etal-2019-top}                  & 73.2 \\
                         & \citet{P18-1037}                           & 74.4$\pm$0.2                  \\ 
                         & \citet{lindemann-etal-2019-compositional}  & 75.3$\pm$0.1 \\
                         & \citet{naseem-etal-2019-rewarding}         & 75.5 \\
                         & \citet{sheng-etal-2019-amr}                           & 76.3$\pm$0.1                  \\ 
                         & ~~~~- w/o beam search                           & 75.3$\pm$0.1                  \\ \cmidrule(l){2-3}
                         & Ours                       & \textbf{77.0}$\pm$0.1                  \\ 
                         & ~~~~- w/o beam search                           & 76.4$\pm$0.1                  \\ \midrule
\multirow{5}{*}{\specialcell{AMR\\ 1.0}} & \citet{S16-1186}                            & 66.0                  \\
                         &   \citet{D15-1136}                        & 67.1                  \\
                         &   \citet{D17-1129}                         & 68.1                  \\
                         &   \citet{D18-1198}                         & 68.3$\pm$0.4                  \\ 
                         & \citet{sheng-etal-2019-amr}                           & 70.2$\pm$0.1                  \\ 
                         & ~~~~- w/o beam search                           & 69.2$\pm$0.1                  \\ \cmidrule(l){2-3}
                         & Ours                       & \textbf{71.3}$\pm$0.1                  \\ 
                         & ~~~~- w/o beam search                           & 70.4$\pm$0.1                  \\ \midrule
\end{tabular}
\caption{\textsc{Smatch} F1 on AMR 2.0 and 1.0 test sets. Standard deviation is computed over 3 runs.
\label{tab:amr-results}}
\end{table}

\noindent\textbf{AMR} \Cref{tab:amr-results} compares our neural transducer
to the previous best results (\textsc{smatch} F1, \citealp{P13-2131}) on AMR test sets.
The transducer improves the state of the art on AMR 2.0 by 0.7\% F1.
On AMR 1.0 where training data is much smaller than AMR 2.0,
it shows a larger improvement (1.1\% F1) over the state of the art.

In \Cref{tab:amr-results}, we also conduct ablation study 
on beam search to investigate contributions from
the model architecture itself and the beam search algorithm.
The transducer model without beam search is already better than the previous
best parser that is equipped with beam search.
When compared with the previous best parser without beam search,
our model still has around 1.0\% F1 improvement.  

\begin{table}[!ht]
\centering
\begin{tabular}{@{}lcccl@{}}
\toprule
Metric       & L'18 & N'19 & Z'19 &\multicolumn{1}{c}{Ours} \\ \midrule
\textsc{Smatch}  & 74 & 75   & 76  & \textbf{77} \\ \midrule
Unlabeled    & 77  & \textbf{80} & 79 & \textbf{80} \\
No WSD       & 76  & 76 & 77 & \textbf{78} \\
Reentrancies & 52  & 56 & 60 & \textbf{61} \\
Concepts     & \textbf{86} & \textbf{86} & 85   & \textbf{86} \\
Named Ent.   & \textbf{86} & 83 & 78 & 79 \\
Wikification & 76   & 80 & \textbf{86} & \textbf{86} \\
Negation     & 58   & 67 & 75 & \textbf{77} \\
SRL          & 70   & \textbf{72} & 70 & 71 \\ \bottomrule
\end{tabular}
\caption{Fine-grained F1 scores on the AMR 2.0 test set.
L'18 is \citet{P18-1037}; N'19 is \citet{naseem-etal-2019-rewarding}; Z'19 is \citet{sheng-etal-2019-amr}.
\label{tab:amr-individual-phenom}}
\end{table}

\Cref{tab:amr-individual-phenom} summarizes the parser performance on each
subtask using \citet{E17-1051} evaluation tool.
Our transducer outperforms \citet{sheng-etal-2019-amr} on all subtasks, but is still not close to \citet{P18-1037} on named entities
due to the different preprocessing methods for anonymization.

\begin{table}[!ht]
\centering
\begin{tabular}{@{}lll@{}}
\toprule
  Parser        & ID   & OOD  \\ \midrule
\citet{du-etal-2015-peking}        & 89.1 & 81.8 \\
\citet{almeida-martins-2015-lisbon}$^{(\text{open)}}$& 89.4 & 83.8 \\
\citet{AAAI1816549}      & 90.3 & 84.9 \\
\citet{P17-1186}: \textsc{basic} & 89.4 & 84.5 \\
\citet{P17-1186}: \textsc{freda3} & 90.4 & 85.3 \\
\citet{peng-etal-2018-learning}  & 91.2 & 86.6 \\
\citet{P18-2077} & \textbf{93.7} & \textbf{88.9} \\ \midrule
Ours      & 92.2 &  87.1    \\ \bottomrule
\end{tabular}
\caption{Labeled F1 (\%) scores on the English DM in-domain (WSJ) and out-of-domain (Brown corpus) test sets.
$^{(\text{open)}}$ denotes results from the open track.}
\label{tab:dm-results}
\end{table}

\noindent\textbf{DM} \Cref{tab:dm-results} compares our neural transducer
to the state of the art (labeled F1) on the English DM in-domain (ID) and
out-of-domain (OOD) data.
Except~\citet{P18-2077}, our transducer outperforms all other baselines,
including \textsc{freda3} of \citet{P17-1186} and \citet{peng-etal-2018-learning}, 
which leverage multi-task learning from different datasets. 
The best parser \citep{P18-2077} is specifically designed for
bi-lexical dependencies, and is not directly applicable to 
other semantic parsing tasks such as AMR and UCCA.
In contrast, our transducer is more general, and is competitive 
to the best SDP parser.

\begin{table}[!ht]
\centering
\begin{tabular}{@{}ll@{}}
\toprule
 Parser    & F1 (\%) \\ \midrule
\citet{hershcovich-etal-2017-transition}    & 71.1    \\
\citet{hershcovich-etal-2018-multitask}: single    & 71.2    \\
\citet{hershcovich-etal-2018-multitask}: MTL    & 74.3    \\ 
\citet{jiang-etal-2019-hlt}    & \textbf{80.5}    \\ \midrule
Ours & 76.6$\pm$0.1    \\ 
~~~~- w/o non-terminal node labels & 75.7$\pm$0.1    \\ \bottomrule
\end{tabular}
\caption{Labeled F1 (\%) scores for all edges including primary edges and remote edges.
Standard deviation is computed over 3 runs.
}
\label{tab:ucca-results}
\end{table}

\noindent\textbf{UCCA} \Cref{tab:ucca-results} compares our results
to the previous best published results (labeled F1 for all edges)
on the English Wiki test set.
The best performance is from \citet{jiang-etal-2019-hlt}, where they convert UCCA graphs to constituency trees, and
train a framework for constituency parsing and remote edge recovery.
\citet{hershcovich-etal-2018-multitask} explore multi-task learning (MTL)
to improve UCCA parsing, using AMR, DM and UD parsing as auxiliaries.
While improvement is achieved UCCA parsing, 
their MTL model shows poor results on the auxiliary tasks: 
64.7\% unlabeled F1 on AMR, 27.2\% unlabeled F1 on DM, and 4.9\% UAS on UD.
In comparison, our transducer improves the state of the art on AMR, and shows competitive results on DM.
At the same time, it also shows reasonable results on UCCA.
When converting UCCA DAGs to the unified format, 
we adopt a simple rule (\Cref{sec:unified-format}) to add node labels to non-terminals. 
\Cref{tab:ucca-results} shows that these node labels do improve the parsing performance from 75.7\% to 76.6\%.

\subsection{Analysis}
\label{sec:analysis}

\noindent\textbf{Validity} Graph-based parsers like \citet{P18-2077,sheng-etal-2019-amr}
make independent decisions on edge types. 
As a result, the same outgoing edge type can appear multiple times to a node.
For instance, a node can have more than one \texttt{ARG1} outgoing edge.
Although F1 scores can be computed for graphs with such kind of nodes,
these graphs are in fact invalid mean representations.
Our neural transducer incrementally builds meaning representations: at each decoding step, it takes a semantic relation as input, and has memory of preceding edge type information, 
which implicitly places constraints on edge type prediction.   
We compute the number of invalid graphs predicted by the parser of \citet{sheng-etal-2019-amr} and our neural transducer
on the AMR 2.0 test set,
and find that our neural transducer reduces the number of invalid graphs by 8\%.

\noindent\textbf{Speed}
Besides the improvement on parsing accuracy, we also significantly speed up
parsing.
\Cref{tab:speed} compares the parsing speed of our transducer and \citet{sheng-etal-2019-amr}
on the AMR 2.0 test set, under the same environment setup.
Without relying on MST algorithms to produce 
a valid arborescence, our transducer is able to parse at 1.7x speed. 

\begin{table}[!ht]
\centering
\begin{tabular}{@{}lc@{}}
\toprule
     & Speed (tokens/sec) \\ \midrule
\citet{sheng-etal-2019-amr}    & 617               \\
Ours & \textbf{1076}                \\ \bottomrule
\end{tabular}
\caption{Parsing speed on the AMR 2.0 test set.}
\label{tab:speed}
\end{table}

\section{Conclusion}
We cast three broad-coverage semantic parsing tasks into a unified transduction framework, and propose a
neural transducer to tackle the problem.
Given the input text, 
the transducer incrementally builds a meaning representation via a sequence of semantic relations.
Experiments conducted on three tasks show that our approach improves the state of the art
in both AMR and UCCA, and is competitive to the best parser in SDP.

This work can be viewed as a starting point for cross-framework semantic parsing.
Also, compared with transition-based parsers (e.g. \citealp{E17-1051}) and graph-based parsers (e.g. \citealp{P18-2077}), our transductive framework does not require a pre-trained aligner, and it is capable of building a meaning representation that is less anchored to the input text. These advantages make it well suited to semantic parsing in cross-lingual settings~\cite{x-dsp}.
In the future, we hope to explore its potential in cross-framework and cross-lingual semantic parsing.

\section*{Acknowledgments}
We thank the anonymous reviewers for their
valuable feedback. 
This work was supported in part by the
JHU Human Language Technology Center of Excellence, and DARPA LORELEI and
AIDA. The U.S. Government is authorized to reproduce and distribute reprints for Governmental
purposes. The views and conclusions contained in
this publication are those of the authors and should
not be interpreted as representing official policies
or endorsements of DARPA or the U.S. Government.

\bibliography{emnlp-ijcnlp-2019}
\bibliographystyle{acl_natbib}

\end{document}


\maketitle

\appendix

\section{Appendices}
\label{sec:appendix}

\subsection{Speed}

\begin{table}[!ht]
\centering
\begin{tabular}{@{}lc@{}}
\toprule
     & Speed (tokens/sec) \\ \midrule
\citet{sheng-etal-2019-amr}    & 617               \\
Ours & \textbf{1076}                \\ \bottomrule
\end{tabular}
\caption{Parsing speed on the AMR 2.0 test set.}
\label{tab:speed}
\end{table}

 Our attention-based neural transducer extends the two-stage graph-based parser of \citet{sheng-etal-2019-amr}.
Beside the improvement on parsing accuracy, we also significantly improve
the parsing speed.
\Cref{tab:speed} compares the parsing speed of our transducer and \citet{sheng-etal-2019-amr}
on the AMR 2.0 test set, under the same environment setup.
Without relying on MST algorithms~\citep{chu-liu-1965,edmonds1968optimum} to produce 
a valid arborescence, our transducer is able to parse at 1.7x speed. 

\subsection{Decoder Functions}
The feed-forward neural networks(\textsc{ffn}), multi-layer perceptrons (\textsc{MLP}), 
\textsc{biaffine} and \textsc{bilinear} functions in the decoder are defined as below:
\begin{gather*}
\textsc{ffn}(\mathbf{x}) = \mathbf{W}\mathbf{x}+\mathbf{b}\\
\textsc{mlp}(\mathbf{x}) = \textsc{elu}(\mathbf{W}\mathbf{x}+\mathbf{b})\\
\textsc{biaffine}(\mathbf{x}_1, \mathbf{x}_2) = \mathbf{x}^\top_1\mathbf{U}\mathbf{x}_2 + \mathbf{W}[\mathbf{x}_1;\mathbf{x}_2] + \mathbf{b} \\
\textsc{bilinear}(\mathbf{x}_1, \mathbf{x}_2) = \mathbf{x}^\top_1\mathbf{U}\mathbf{x}_2 + \mathbf{b} 
\end{gather*}
where $\mathbf{U}, \mathbf{W}, \mathbf{b}$ are learned parameters;
\textsc{elu} is an exponential linear unit~\citep{clevert2015fast}.

\subsection{Beam Search over Relations}
At test time, we implement beam search over semantic relations.
We explain it in Algorithm \ref{algo:beamsearch}.

\begin{algorithm}[!ht]
\SetAlCapNameFnt{\small}
\SetAlCapFnt{\small}
\small
\SetKwData{Bucket}{\textsc{bucket}}
\SetKwData{Beam}{beam}
\SetKwData{NewBeam}{new\_beam}
\SetKwInOut{Input}{Input}
\SetKwInOut{Output}{Output}

\Input{The input text $X$.}
\Output{A sequence of relations $Y=\{y_1,...y_m\}$.}
\tcp{Initialization.}
$i\leftarrow 0$\;
$Y \leftarrow \{\}$\;
$\text{score} \leftarrow 0$\;
\texttt{finished}$\leftarrow \{\}$\;
\texttt{beam}$\leftarrow \{ \{Y, \text{score}\}\}$\;
\BlankLine
\tcp{Encoding.}
\texttt{encode($X$)}\;
\BlankLine
\tcp{Decoding.}
\For{$i\leftarrow 1$ \KwTo \text{\upshape MaxLength}}{
    \texttt{new\_beam} $\leftarrow \{\}$\;
    $\{Y, \text{score}\}$ = \texttt{beam.pop()}\;
    \tcp{Run beam search.}
    \For{$v_i$ \text{\upshape in} \texttt{\upshape topK}($\text{\upshape P}(v_i)$)}{
        \uIf{$v_i$ = \texttt{\upshape EOS}}{
            \texttt{finished.push}($\{Y, \text{score}\})$\;
        }\Else{
            \For{$u_i\leftarrow v_0$ \KwTo $v_{i-1}$}{
                \For{$r_i$ \text{\upshape in RelationTypeSet}}{
                    $Y \leftarrow Y\cup\{\langle u_i, r_i, v_i\rangle\}$\;
                    $\text{score} \leftarrow \text{score} + \log\big(\text{P}(u_i)\big) + \log\big(\text{P}(r_i)\big) + \log\big(\text{P}(v_i)\big)$\;
                    \texttt{new\_beam.push}(\{Y, \text{score}\})\;
                }
            }
        }
    }
    \texttt{beam} $\leftarrow$ \texttt{new\_beam.topK}()\; 
}
\BlankLine
\tcp{Finishing.}
\While{\texttt{\upshape beam.not\_empty()}}{
    $\{Y, \text{score}\}$ $\leftarrow$ \texttt{beam.pop()}\;
    \texttt{finished.push}($\{Y, \text{score}\})$\;
}
$\{Y, \text{score}\}$ $\leftarrow$ \texttt{finished.topK(k=1)}\;
\Return $Y$\;
\caption{Beam Search over Semantic Relations.\label{algo:beamsearch}}
\end{algorithm}

\subsection{Hyper-parameter Settings}
\Cref{tab:hyper} lists the hyperparameters of our model for each task.

\begin{table}[!ht]
\centering
\scalebox{0.8}{
\begin{tabular}{ l c }
\multicolumn{2}{c}{\textbf{Hidden Size}}\\ 
Glove                                      &300              \\ 
BERT                                     &1024              \\ 
                                   &100              \\ 
POS / NER / Dep / Shapes &100 \\
\noindent\begin{tabular}{@{}l}
Anonymization / Node index\\
\end{tabular} &50 \\ 
CharCNN kernel size &3 \\
CharCNN channel size &100 \\
Encoder &2@512 \\
Decoder &2@1024 \\
Biaffine input size  &256                                \\
Bilinear input size  &\begin{tabular}{lc}AMR &128 \\ DM & 256   \\ UCCA&128  \end{tabular}                                  \\
\multicolumn{2}{c}{\textbf{Optimizer}} \\
Type                                    &ADAM             \\
Learning rate                          &0.001           \\
Maximum gradient norm                         &5.0              \\
Coverage loss weight $\lambda$ &1.0              \\
Label smoothing $\epsilon$     &0.1              \\
Beam size                      &5                \\
Batch size                     &64           \\
Dropout rate &\begin{tabular}{lc}AMR &0.33 \\ DM & 0.2   \\ UCCA&0.33  \end{tabular} \\
\multicolumn{2}{c}{\textbf{Vocabulary}}                                                    \\
Encoder-side vocab size                 & \begin{tabular}{lc}AMR 1.0 &9200 \\AMR 2.0 &18000   \\ DM & 11000   \\ UCCA&10000  \end{tabular}  \\
Decoder-side vocab size& \begin{tabular}{lc}AMR 1.0 &7300 \\AMR 2.0 &12200    \\ DM & 11000   \\ UCCA&10000  \end{tabular}  \\
\end{tabular}
}

\caption{Hyper-parameter settings}
\label{tab:hyper}
\end{table}

\bibliography{emnlp-ijcnlp-2019}
\bibliographystyle{acl_natbib}